\documentclass[letterpaper]{article} 
\usepackage{aaai23}  
\usepackage{times}  
\usepackage{helvet}  
\usepackage{courier}  
\usepackage[hyphens]{url}  
\usepackage{graphicx} 
\usepackage{natbib}  
\usepackage{caption} 
\usepackage{algorithm}
\usepackage{algorithmic}

\usepackage{newfloat}
\usepackage{listings}
\DeclareCaptionStyle{ruled}{labelfont=normalfont,labelsep=colon,strut=off} 
\floatstyle{ruled}
\newfloat{listing}{tb}{lst}{}
\floatname{listing}{Listing}
\usepackage{booktabs}
\usepackage{amsmath}
\usepackage{amssymb}
\usepackage{amsfonts}
\usepackage{makecell}
\usepackage[colorlinks,
linkcolor=blue,
citecolor=blue
]{hyperref}
\usepackage{color}

\title{Semi-supervised Deep Large-baseline Homography Estimation with Progressive Equivalence Constraint}
\author{
	Hai Jiang\textsuperscript{\rm 1,\rm 3}\equalcontrib, Haipeng Li\textsuperscript{\rm 2,\rm 3}\equalcontrib, Yuhang Lu\textsuperscript{\rm 4}, Songchen Han\textsuperscript{\rm 1,\footnotemark[2]}, Shuaicheng Liu\textsuperscript{\rm 2,\rm 3,\footnote{Corresponding authors}}
}
\affiliations{
	\textsuperscript{\rm 1}Sichuan University\\
	\textsuperscript{\rm 2}University of Electronic Science and Technology of China\\
	\textsuperscript{\rm 3}Megvii Technology\\
	\textsuperscript{\rm 4}Univesity of South Carolina\\
	\{jianghai1@stu., hansongchen@\}scu.edu.cn, \{lihaipeng@stu.,liushuaicheng@\}@uestc.edu.cn\\
	\{jianghai,lihaipeng,liushuaicheng\}@megvii.com, yuhang@email.sc.edu
}

\usepackage{bibentry}

\begin{document}
	\maketitle
	\begin{abstract}
		Homography estimation is erroneous in the case of large-baseline due to the low image overlay and limited receptive field. To address it, we propose a progressive estimation strategy by converting large-baseline homography into multiple intermediate ones, cumulatively multiplying these intermediate items can reconstruct the initial homography. Meanwhile, a semi-supervised homography identity loss, which consists of two components: a supervised objective and an unsupervised objective, is introduced. The first supervised loss is acting to optimize intermediate homographies, while the second unsupervised one helps to estimate a large-baseline homography without photometric losses. To validate our method, we propose a large-scale dataset that covers regular and challenging scenes. Experiments show that our method achieves state-of-the-art performance in large-baseline scenes while keeping competitive performance in small-baseline scenes. Code and dataset are available at https://github.com/megvii-research/LBHomo.
	\end{abstract}
	\section{Introduction}
	Homography estimation is a basic task in computer vision that has been widely used for a wide range of high-level vision tasks, such as image/video stitching~\cite{image/video1}, video stabilization~\cite{stabilization}, SLAM~\cite{SLAM}, and HDR reconstruction~\cite{HDR1,HDR2}. Traditional methods typically use feature detection and matching algorithms~\cite{sift,orb}, and subsequently solve direct linear transform ($\mathcal{DLT}$)~\cite{DLT} with outlier suppression to obtain a homography matrix. However, these methods are highly dependent on the extracted feature matches and may fail in scenes that lack sufficient high-quality feature points.
	\begin{figure}[!ht]
		\centering
		\includegraphics[width=\linewidth]{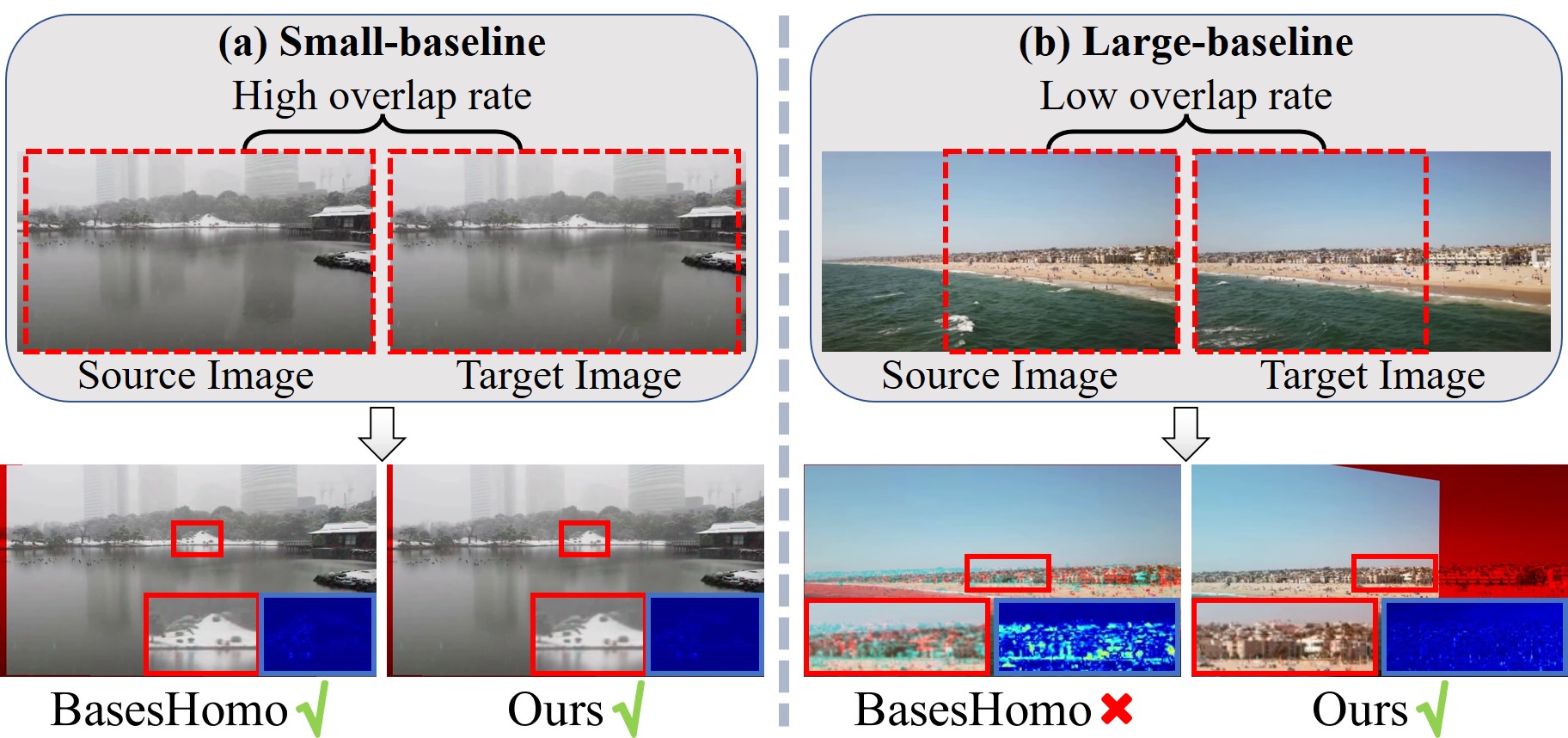}
		\caption{Comparison of our method with an unsupervised learning-based method, i.e., BasesHomo. The first row shows two pairs of consecutive frames in different baseline cases, we highlight the overlap region in red boxes. In the second row, we superimpose the warped source image and the target image, where misaligned pixels are visualized as colored ghosts. Local details are enlarged in red and blue boxes, where the latter is an error heatmap, the darker the better. Our method can handle both situations, while BasesHomo fails in the large-baseline scene.}
		\label{fig:figure1compare}
	\end{figure}
	On the other hand, learning-based methods take a pair of source and target images as input and directly output the corresponding homography matrix. They do not rely on matched key points and thus are more robust than traditional methods. The learning-based methods can be divided into two categories: supervised methods~\cite{supervised2016} and unsupervised methods~\cite{unsupervised2018}. The supervised ones use synthetic image pairs for training due to the lack of sufficient real-world image pairs with GT labels, lacking realistic scene parallax causes unsatisfactory generalization ability. Unsupervised methods adopt label-free training strategies that aim to minimize the photometric distance between the warped source images and target images, being better generalized to various scenes. With the assistance of photometric losses~\cite{CA-Unsupervised2020,BasesHomo2021}, the unsupervised methods perform well in small-baseline scenes where the non-overlap rate between two images is less than 10$\%$. However, in large-baseline scenes where the non-overlap rate is between 20$\%$ and 50$\%$, the warped source image contains a number of out-of-boundary pixels due to the large appearance and viewpoint changes, causing it hard to minimize the photometric distance. As shown in Fig.~\ref{fig:figure1compare}, BasesHomo~\cite{BasesHomo2021} can successfully align two images with large overlap but fails in small overlap, while our method is capable to handle both cases. To address large-baseline cases,~\cite{stitching1} proposed an ablation-based strategy, which forces the two images to contain the same size of valid areas to ignore out-of-boundary pixels. However, this strategy fails where dynamic objects exist in images. From the above, we find it is non-trivial to estimate the homography of two images with a large-baseline. 
	\begin{figure*}[!ht]
		\centering
		\includegraphics[width=0.98\linewidth]{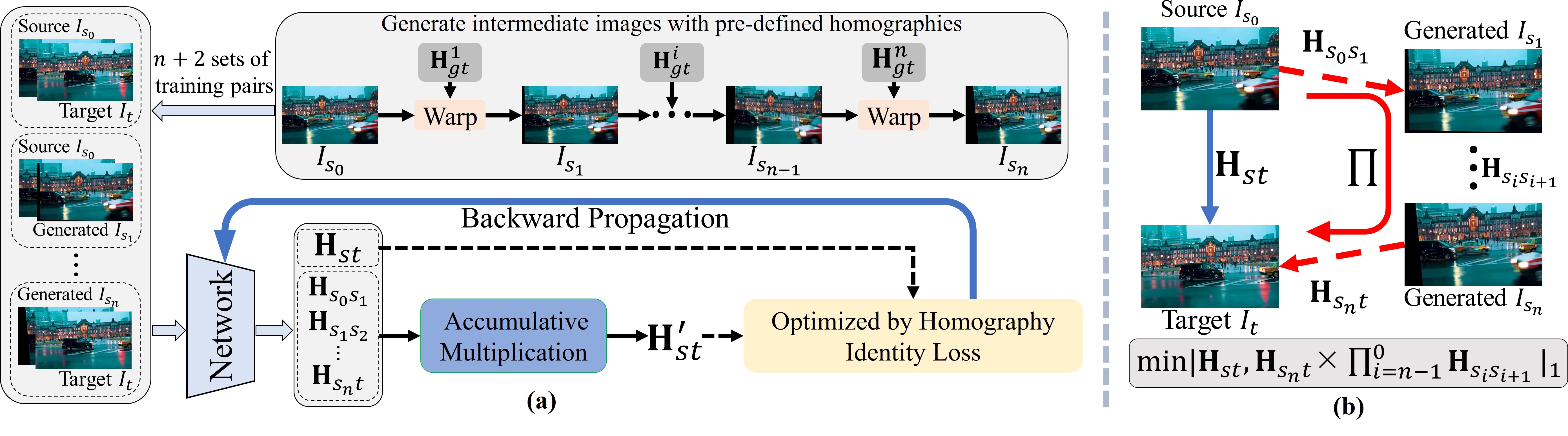}
		\caption{The left column (a) illustrates our proposed progressive estimation strategy, and right column (b) illustrates our proposed unsupervised objective function. $\prod$ denotes the accumulative multiplication operation.}
		\label{fig:progressive strategy}
	\end{figure*}
	To this end, we propose a progressive homography estimation strategy to address the large-baseline challenge. Specifically, we convert the large-baseline problem into multiple intermediate phases by inserting several images along with pre-defined homographies into the source and target image, as shown in Fig.~\ref{fig:progressive strategy}(a). The homography between the source image and the target image can be obtained by cumulatively multiplying these intermediate homographies. To achieve this, we design a homography identity loss to optimize our network in a semi-supervised way by considering the advantages and weaknesses of supervised and unsupervised learning. Our loss function consists of two parts: a supervised objective and an unsupervised objective. Firstly, given the pre-defined homographies, the supervised one is utilized to optimize intermediate homographies of inserted images. Secondly, the unsupervised loss helps to estimate a large-baseline homography without photometric loss. More specifically, given the target image, our network predicts two homographies, the former being the last intermediate image to it and the later being the source towards it. Our unsupervised loss minimizes the error between the homography of the source and target image and the accumulative multiplication result of all intermediate items, i.e. the homographies of inserted images and the homography of the last intermediate and target image, as shown in Fig.~\ref{fig:progressive strategy}(b). 
	
	With our progressive estimation strategy and semi-supervised loss, the network can focus on homography optimization while avoiding the problems caused by photometric losses. Additionally, we introduce a large-scale real-world image pairs dataset for large-baseline homography estimation considering there lacks a dedicated dataset for such a task, which contains 5 categories of scenes as well as human-labeled GT point correspondences for quantitative evaluation. Extensive experiments demonstrate that our method outperforms the state-of-the-art supervised methods and unsupervised methods both quantitatively and qualitatively. Moreover, our method is also applicable to small-baseline scenarios. To summarize, our main contributions are threefold:
	\begin{itemize}
		\item We propose a progressive estimation strategy to address large-baseline homography estimation by transforming the large-baseline into several intermediate ones.
		\item We propose a semi-supervised homography identity loss that enforces the network to focus on optimizing the homography.
		\item We introduce a large-scale dataset containing various scenes for large-baseline homography learning and the human-labeled evaluation set is also included. Experimental results demonstrated that our method achieves state-of-the-art performance.
	\end{itemize}
	
	\section{Related Work}
	\subsection{Traditional Homography Estimation}
	Traditional homography estimation methods usually combine classic or learning-based feature extraction and matching algorithms such as SIFT~\cite{sift}, ORB~\cite{orb}, BEBLID~\cite{beblid}, SuperPoint~\cite{superpoint}, SOSNet~\cite{sosnet}, SuperGlue~\cite{superglue}, LoFTR~\cite{loftr}, and subsequently solve direct linear transform with outliner suppression such as RANSAC~\cite{ransac}, MAGSAC~\cite{magsac}. However, feature-based methods usually crash in challenging scenes where sufficient feature matches cannot be obtained. In addition, some methods can also solve a homography directly by using the Lucas-Kanade algorithm~\cite{lucas} or calculating the sum of squared differences (SSD) between two images without extracted feature matches. A randomly initialized homography is optimized in this way iteratively.
	
	\subsection{Deep Homography Estimation}
	Following the development of learning-based image alignment methods, such as optical flow~\cite{GyroFlow,UPFlow,realflow} and dense correspondence~\cite{PDC-Net,DeepOIS}, a deep homography estimation network was first proposed by~\cite{supervised2016}. Deep homography estimation methods can be divided into two categories: supervised and unsupervised. The former ones~\cite{Crossresolution-supervised2021,Iterative-supervised2022} utilize the generated image pairs with ground-truth labels to train their models, but their generalization ability is limited due to the lack of realistic scene parallax in synthetic images. Unsupervised methods~\cite{unsupervised2018,unsupervised2020,BasesHomo2021} optimize their models with real-world image pairs by minimizing the photometric distance from the source image warped by the estimated homography to the target image. To be more robust, some methods~\cite{CA-Unsupervised2020,HomoGAN2022} introduce efficient masks to replace classic outlier rejection methods to remove undesired regions or focus on the dominant plane. However, most of the previous methods are proposed to estimate the homography of image pairs with a small-baseline, large-baseline homography estimation, a field with broader applications, has long been ignored.
	
	\subsection{Deep Image Alignment}
	Recently, some image stitching methods~\cite{stitching1,stitching3} use an individual homography estimation network for coarse alignment and optimize the pre-aligned images by reconstruction networks to achieve better stitching results in large-baseline scenes. However, their networks are not specially designed for such tasks, leading to unsatisfactory results. In addition, some geometric matching methods~\cite{DAMG,GLU-Net,warpc} can also be applied to solve the large-baseline image alignment. But their alignment is mainly realized by mesh flow or dense flow, which contains the local motion information of the images, while the homography matrix only represents the global motion. In contrast, we propose a progressive estimation strategy and a semi-supervised consistency constraint without photometric loss to address the large-baseline homography estimation.
	
	\section{Method}
	\subsection{Overview}
	In this section, we introduce a progressive strategy to address large-baseline homography estimation. Specifically, we transform the large-baseline into several intermediate stages by inserting some intermediate images, i.e., $I_{s_{i}} \in \mathbb{R}^{H \times W \times 3}$ ($i \in [0, n]$), into the source image $I_s \in \mathbb{R}^{H \times W \times 3}$ and target image $I_t \in \mathbb{R}^{H \times W \times 3}$. We generate the intermediate ones by a set of pre-defined homographies, i.e., $I_{s_{i+1}}=\mathcal{W}_{i+1}(I_{s_{i}}, \textbf{H}_{gt}^{i+1})$, where $\mathcal{W}_{i+1}$ represents the warp operation by the $\textbf{H}_{gt}^{i+1}$ and $I_{s_{0}}$ is the initial $I_{s}$. By randomly sampling the pre-defined homographies with non-identity matrices, it is effective to avoid the degenerate solutions of accumulative multiplication results of intermediate homographies~\cite{warpc}. In addition, we ensure that the non-overlap rate of the two inserted images is smaller than that of the source and target image.
	
	After generating $n$ intermediate images, we get $n+2$ sets of image pairs, i.e., $(I_{s_{i}}, I_{s_{i+1}})$, $(I_{s_{n}}, I_{t})$, and $(I_{s}, I_{t})$. Our goal is to train a neural network $f_{\theta}$, with parameters $\theta$, that predicts homography matrix $\textbf{H}_{s_{i}s_{i+1}} = f_{\theta}(I_{s_{i}}, I_{s_{i+1}})$, $\textbf{H}_{s_{n}t} = f_{\theta}(I_{s_{n}}, I_t)$, $\textbf{H}_{st} = f_{\theta}(I_s, I_t)$ relating $I_{s_{i}}$ to $ I_{s_{i+1}}$, $I_{s_{n}}$ to $ I_t$, and $I_s$ to $ I_t$ respectively. Multiplying all of the intermediate homographies $\textbf{H}_{s_{i}s_{i+1}}$ and the $\textbf{H}_{s_{n}t}$ should be equal to the $\textbf{H}_{st}$. With this equivalence constraint, we can enforce the network to optimize the homographies themselves. To achieve this, we propose a semi-supervised homography identity loss to train our network, which is described in the following section.
	\begin{figure*}[!ht]
		\centering
		\includegraphics[width=0.98\linewidth]{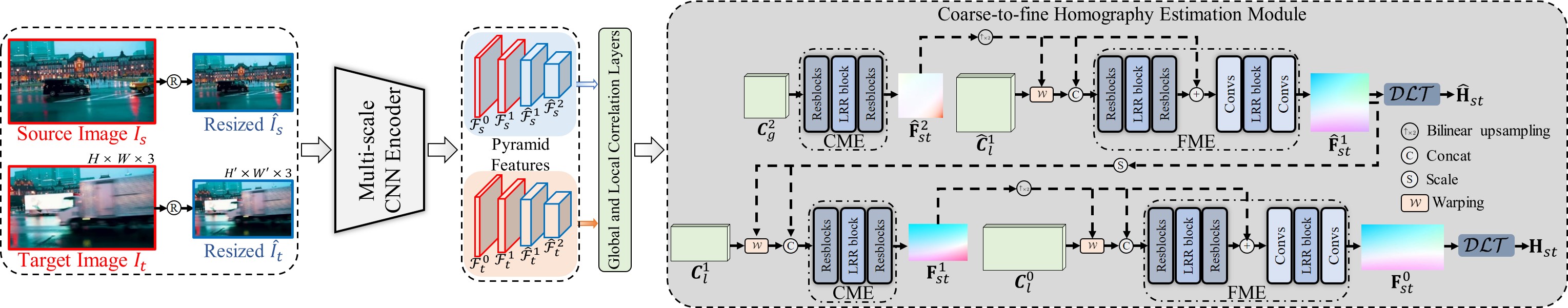}
		\caption{The overall pipeline of our proposed network. Firstly, we resize the original images into lower resolution ones and use a multi-scale CNN encoder to obtain the pyramid features. The correlation layers are used to improve the utilization of feature information and expand the receptive field. Finally, we use a coarse-to-fine homography estimation module to obtain homography flows and solve $\mathcal{DLT}$ to get the corresponding homography matrices.}
		\label{fig:network architecture}
	\end{figure*}
	\subsection{Homography Identity Loss}
	The point-wise correspondence between each set of image pairs can be the mapping of the corresponding homography matrix. Let us denote $\textbf{X}_{s_{i}}$ and $\textbf{X}_t$ as the meshgrid coordinate sets of $I_{s_{i}}$ and $I_t$, respectively. The point-wise correspondence of $\textbf{X}_{s_{i}}$ and $\textbf{X}_{s_{i+1}}$ is associated by the $\textbf{H}_{s_{i}s_{i+1}}$ using $\textbf{X}_{s_{i+1}} = \textbf{H}_{s_{i}s_{i+1}} \textbf{X}_{s_{i}}$. Likewise, the correspondence of $\textbf{X}_{s_{n}}$ and $\textbf{X}_t$ is related by the $\textbf{H}_{s_{n}t}$. Accordingly, the coordinate correspondence between the source and target image can be expressed as $\textbf{X}_t = \textbf{H}_{st} \textbf{X}_s = (\textbf{H}_{s_{n}t} \times \textbf{H}_{s_{n-1}s_{n}} \times \cdots \times \textbf{H}_{s_{0}s_{1}}) \textbf{X}_s$, where $\times$ denotes the cross-product operation. Therefore, the homography of the source image and the target image $\textbf{H}_{st}$ can be obtained by multiplying the $\textbf{H}_{s_{n}t}$ and $\textbf{H}_{s_{i}s_{i+1}}$. The corresponding identity equation can be expressed as
	\begin{equation}\small
		\prod \limits_{i=n-1}^0 \textbf{H}_{s_{i}s_{i+1}} = \textbf{H}_{s_{n}t}^{-1} \times \textbf{H}_{st}.
		\label{equation:eq3}
	\end{equation}
	In essence, our optimization goal is not to minimize the distance between the warped source image and the target image, but to minimize the error between estimated homographies based on Eq.(\ref{equation:eq3}) in an unsupervised manner. The unsupervised objective function ($\mathcal{L}_{unsup}$) is formulated as 
	\begin{equation}\small
		\mathcal{L}_{unsup} = |\textbf{H}_{s_{n}t}^{-1} \times \textbf{H}_{st} - \prod \limits_{i=n-1}^0 \textbf{H}_{s_{i}s_{i+1}}|_1,
		\label{equation:eq4}
	\end{equation}
	where $|\cdot|_1$ denotes the L1 norm. Since the intermediate images are generated through the pre-defined homographies, so we can estimate the homographies of intermediate images in a supervised way. The supervised objective function ($\mathcal{L}_{sup}$) is formulated as 
	\begin{equation}\small
		\mathcal{L}_{sup} = \sum \limits_{i=0}^{n-1} |\textbf{H}_{s_{i}s_{i+1}} - \textbf{H}_{gt}^{i+1}|_1,
		\label{equation:eq5}
	\end{equation}
	and the $\textbf{H}_{s_{i}s_{i+1}}$ in Eq.(\ref{equation:eq4}) can be replaced with the $\textbf{H}_{gt}^{i+1}$. However, due to the cancellation effect amid the estimated homography terms, artlessly replacing $\textbf{H}_{s_{i}s_{i+1}}$ with $\textbf{H}_{gt}^{i+1}$ may obtain degeneration solutions, e.g., $\textbf{H}_{st} = I_{3\times3}$. To avoid this situation, we rewrite Eq.(\ref{equation:eq4}) as
	\begin{equation}\small
		\mathcal{L}_{unsup} =  \sum \limits_{i=0}^{n-1} \lambda_{i} |(\textbf{H}_{s_{i+1}t}^{-1} \times \textbf{H}_{st}) -  \prod \limits_{j=i}^0 \textbf{H}_{gt}^{j+1}|_1,
		\label{equation:eq6}
	\end{equation}
	the detailed derivation is conducted in the supplementary material.
	
	Our final semi-supervised homography identity loss $\mathcal{L}_{HIL}$ combines the $\mathcal{L}_{sup}$ and $\mathcal{L}_{unsup}$ as $\mathcal{L}_{HIL} = \mathcal{L}_{unsup} + \lambda_{w} \mathcal{L}_{sup}$, where the $\lambda_{w}$ is a weighting factor, we eliminate this hyper-parameter by automatically balancing the weights over each training batch as $\lambda_{w}=\mathcal{L}_{unsup}/\mathcal{L}_{sup}$.
	
	\subsection{Multi-scale Homography Estimation Network}
	There are two challenges for the large-baseline homography estimation: 1) the overlap rate between two images is low, and 2) the receptive field of CNN-based models is limited~\cite{stitching2}. To overcome these problems, we design a multi-scale homography estimation network, as shown in Fig.~\ref{fig:network architecture}, that combines a multi-scale CNN encoder and correlation layers to leverage feature information and expand the receptive field. Subsequently, the global and local correlation maps are fed into a coarse-to-fine homography estimation module to obtain the final results. 
	
	\noindent \textbf{Multi-scale CNN Encoder.} Given a pair of images $I_{s}$ and $I_{t}$, we use a multi-scale CNN encoder consists of four cascaded residual blocks and max-pooling layers to extract multi-scale features as $\mathcal{F}^{k} \in \mathbb{R}^{\frac{H}{2^{2+k}} \times \frac{W}{2^{2+k}} \times d^{k}}$, $k \in [0, 2]$. Features at the top pyramid layer have the lowest resolution, representing the most global information, which are subsequently used to generate the global correlation map. The size of the global correlation map is determined by the size of the input features, which requires a significant computation in the case of high-resolution images. Therefore, we resize the $I_{s}$ and $I_{t}$ to $\hat{I}_{s}$ and $\hat{I}_{t}$ with fixed lower resolution $H^{'} \times W^{'}$. The $\hat{I}_{s}$ and $\hat{I}_{t}$ are fed into the encoder to produce features as $\hat{\mathcal{F}}^{k}$. As shown in Fig.~\ref{fig:network architecture}, we select $\mathcal{F}^{0}$, $\mathcal{F}^{1}$, $\hat{\mathcal{F}}^{1}$, and $\hat{\mathcal{F}}^{2}$ to form a four-layer feature pyramid. A set of homographies can be estimated from the pyramid layers, which are transmitted in a coarse-to-fine manner.
	
	\noindent \textbf{Feature Correlation.} Previous unsupervised methods~\cite{CA-Unsupervised2020,BasesHomo2021,HomoGAN2022} estimate homography without using correlation layers. However, we find that the correlation is effective for feature matching. Specifically, the $\hat{\mathcal{F}}^{2}$ contains more global information than others, we use the global correlation layer to represents the pairwise similarity between spatial positions in the source feature $\hat{\mathcal{F}}^{2}_{s}$ and target feature $\hat{\mathcal{F}}^{2}_{t}$ as
	\begin{equation}\small
		\textbf{C}_{g}^{2}(\textbf{x}_s, \textbf{x}_t)) = 
		\hat{\mathcal{F}}^{2}_{s}(\textbf{x}_s)^\top \hat{\mathcal{F}}^{2}_{t}(\textbf{x}_t),
		\label{equation:eq8}
	\end{equation}
	where $\textbf{x}_s$ and $\textbf{x}_t$ denote the coordinate position of the source and target feature, and $\textbf{C}_{g}^{2}$ is the global correlation map of the source and target feature. The result is a 4D tensor, we reshape it to a 3D tensor of size $\frac{H^{'}}{16} \times \frac{W^{'}}{16} \times \frac{H^{'} \times W^{'}}{256}$.
	
	For the rest feature maps, i.e., $\mathcal{F}^{0}$, $\mathcal{F}^{1}$, $\hat{\mathcal{F}}^{1}$, we apply the local correlation layer proposed in~\cite{GOCor} to evaluate the feature similarity between two feature maps, denoted as $\textbf{C}_{l}^{0}$, $\textbf{C}_{l}^{1}$, and $\hat{\textbf{C}}_{l}^{1}$. The  search region $R$ is set to constrain the search space and result in local correlation maps with the size of $\textbf{C}_{l}$ is $\frac{H^{(')}}{2^{2+k}} \times \frac{W^{(')}}{2^{2+k}} \times (2R+1)^{2}$.
	
	\noindent \textbf{Coarse-to-fine Homography Estimation Module.} Given feature correlations, we adopt two coarse motion estimators (CME) and two fine motion estimators (FME) to extract the global and local motion from the correlations. As discussed above, the relative motion between $I_{s}$ and $I_{t}$ can be mapped by a homography. Therefore, We compute the homography flow~\cite{GyroFlow} as $\textbf{F}_{st} = \textbf{X}_{t} - \textbf{X}_{s}$ to facilitate the learning of motion information and subsequently solve the $\mathcal{DLT}$ to obtain a unique homography. The homography flow of $\hat{I}_{s}$ and $\hat{I}_{t}$ is expressed as $\hat{\textbf{F}}_{st}$. Specifically, the $\textbf{C}_{g}^{2}$ is fed into the CME to estimate the coarse homography flow $\hat{\textbf{F}}_{st}^{2}$, and the fine $\hat{\textbf{F}}_{st}^{1}$ is obtained by combining $\hat{\textbf{F}}_{st}^{2}$ and $\hat{\textbf{C}}_{l}^{1}$ as input through the FME. Likewise, the homography flow $\textbf{F}_{st}^{1}$ and $\textbf{F}_{st}^{0}$ can be generated with the corresponding features and correlations via motion estimators. Moreover, the LRR blocks~\cite{BasesHomo2021} are inserted at FME to reject motion outliers implicitly, detailed architectures of the CME and FME are illustrated in Fig~\ref{fig:network architecture}. Finally, we convert the flows into the corresponding homographies by solving the $\mathcal{DLT}$. 
	\begin{figure*}[!ht]
		\centering
		\includegraphics[width=0.98\linewidth]{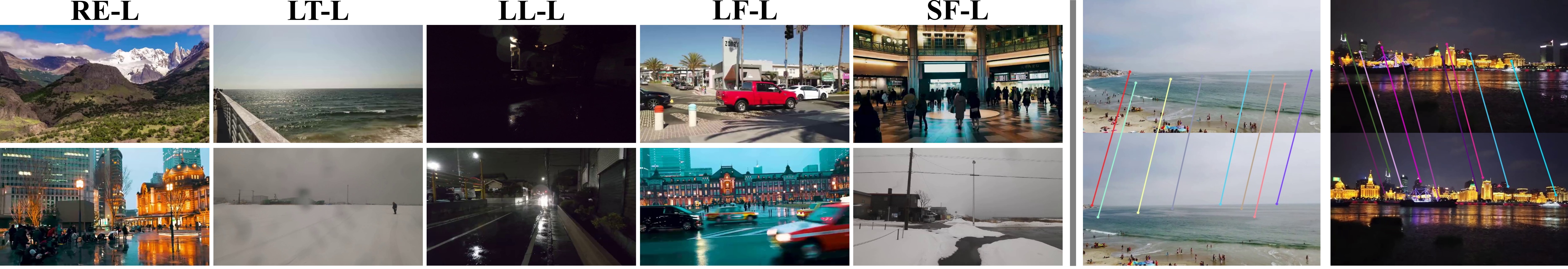}
		\caption{A glace of our dataset. The dataset contains five categories as shown in the first five columns. The rightmost two columns show several examples of human labeled point correspondences for quantitative evaluation.}
		\label{fig:dataset}
	\end{figure*}
	\subsection{Network Training}
	As mentioned in~\cite{supervised2016}, it is non-trivial to directly estimate a homography matrix. Therefore, we use the homography flow as the supervision object during the training stage. Besides, we resize the initial images to fixed-resolution to improve the applicability to high resolution images, our supervised and unsupervised objectives are accordingly converted into
	\begin{equation}\small
		\mathcal{L}_{sup} = \sum \limits_{i=0}^{n-1} |\textbf{F}_{s_{i}s_{i+1}} - \textbf{F}_{gt}^{i+1}|_1 + \sum \limits_{i=0}^{n-1} |\hat{\textbf{F}}_{s_{i}s_{i+1}} - \hat{\textbf{F}}_{gt}^{i+1}|_1,
		\label{equation:eq10}
	\end{equation}
	\begin{equation}\small
		\begin{aligned}
			\mathcal{L}_{unsup} = \sum \limits_{i=0}^{n-1} \lambda_{i} &(|\textbf{F}_{st} - \textbf{F}_{s_{i+1}t} - \sum\limits_{j=i}^0 \textbf{F}_{gt}^{j+1}|_1 \\
			&+  |\hat{\textbf{F}}_{st} - \hat{\textbf{F}}_{s_{i+1}t} - \sum\limits_{j=i}^0 \hat{\textbf{F}}_{gt}^{j+1}|_1).
		\end{aligned}
		\label{equation:eq11}
	\end{equation} 
	The most relative representations of our homography flow are the 8 bases flow~\cite{BasesHomo2021} and optical flow. While the former performs well in small-baseline scenes, it crashes in large-baselines. Compared to optical flow, with the assistance of our supervised objective and LRR blocks, our homography flows tend to represent the global motion between two images. Importantly, the homography flows are only used to facilitate training, and $\mathcal{DLT}$ will be used to transform the flows into homography matrices.
	
	\section{Experiments}
	\textbf{Dataset.} We introduce a large-scale dataset for large-baseline homography estimation considering there lacks a dedicated dataset for this task. Our dataset contains 5 categories, including regular (RE-L), low-texture (LT-L), low-light (LL-L), small-foregrounds (SF-L), and large-foregrounds (LF-L) scenes. We select image pairs from real-world scenes and ensure the average non-overlap rate between the source and target images is from 20$\%$ to 50$\%$. Our dataset contains $\approx$78k image pairs in totally, and 1.8k image pairs are randomly chosen from all categories as the evaluation data. For each evaluation image pair, we manually labeled more than 6 uniform distributed matching points for quantitative comparisons. Some examples of our dataset are illustrated in Fig.~\ref{fig:dataset}.
	
	\noindent \textbf{Implementation Details.} In the training stage, we randomly crop patches of size 320$\times$480 near the center of the initial images as input, and the resolution of resized images is set to $(H^{'} \times W^{'}) = (256 \times 256)$. The number of inserted images is set to $n=2$, and the non-overlap rate of the two intermediate images is less than 20$\%$. We empirically set the $\lambda_{i}$ in Eq.(\ref{equation:eq11}) to $10^{-i}$. Our network is implemented with PyTorch, and the training is performed on four NVIDIA RTX 2080Ti GPUs. The Adam optimizer~\cite{Adam} is adopted with an initial learning rate of $5 \times 10^{-4}$ for model optimization, and it decays by a factor of 0.8 after every epoch. The batch size is set to 16.
	
	\subsection{Comparison with Existing Methods}
	\noindent \textbf{Quantitative Comparison.} We report the quantitative results of all comparison methods on our large-baseline evaluation set in Table~\ref{tab:large-table}, where rows 3-8 are traditional feature-based methods, rows 9-14 are learned feature-based methods, rows 15-18 are deep homography estimation methods, and rows 19-22 are deep image alignment methods which can be applied to large-baseline scenes. $I_{3\times3}$ in the first row refers to the identity transformation, of which the errors reflect the original distance between point pairs. And we have retrained all deep learning-based methods, except the Supervised~\cite{supervised2016}, on our training set represented by $^{*}$.
	\begin{table*}[!ht]
		\centering
		\resizebox{\linewidth}{!}{
			\begin{tabular}{rlllllll}
				\toprule
				1) & & RE-L & LT-L & LL-L & LF-L & SF-L & Avg-L  \\
				\midrule
				2) & $I_{3\times3}$ & 94.60 (+15428.04$\%$) & 106.48 (+550.86$\%$) & 99.42 (+1036.23$\%$) & 43.57 (+407.81$\%$) & 67.27 (+718.37$\%$) & 82.27 (+698.74$\%$) \\ 
				\midrule
				3) & SIFT+RANSAC & \underline{1.66 (+0.00$\%$)} & 26.47 (+61.80$\%$) & 20.37 (+132.80$\%$) & 9.05 (+5.48$\%$) & \underline{8.22 (+0.00$\%$)} & 13.15 (+27.67$\%$) \\
				4) & SIFT+MAGSAC & 1.74 (+13.28$\%$) & \underline{16.36 (+0.00$\%$)} & 28.74 (+228.46$\%$) & 12.39 (+44.41$\%$) & 12.19 (+48.30$\%$) & 14.29 (+38.74$\%$) \\
				5) & ORB+RANSAC & 5.74 (+677.28$\%$) & 71.91 (+339.55$\%$) & 25.19 (+187.89$\%$) & 19.08 (+122.38$\%$) & 44.36 (+439.66$\%$) & 33.26 (+222.91$\%$) \\
				6) & ORB+MAGSAC & 6.68 (+833.32$\%$) & 74.21 (+353.61$\%$) & 26.52 (+203.09$\%$) & 19.85 (+131.35$\%$) & 44.82 (+445.26$\%$) & 34.42 (+234.17$\%$) \\
				7) & BEBLID+RANSAC & 19.68 (2991.32$\%$) & 90.50 (+453.18$\%$) & 52.32 (+497.94$\%$) & 30.76 (+258.51$\%$) & 61.32 (+645.99$\%$) & 50.92 (+394.37$\%$) \\
				8) & BEBLID+MAGSAC & 21.26 (+3253.60$\%$) & 90.96 (+455.99$\%$) & 53.87 (+515.66$\%$) & 31.99 (+272.84$\%$) & 61.36 (+646.47$\%$) & 51.89 (+403.79$\%$) \\
				\midrule
				9) & SOSNet+RANSAC & 1.86 (+33.20$\%$) & 29.76 (+81.91$\%$) & 18.92 (+116.23$\%$) & 13.59 (+58.39$\%$) & 8.96 (+9.00$\%$) & 14.61 (+41.84$\%$) \\
				10) & SOSNet+MAGSAC & 2.02 (+59.76$\%$) & 35.42 (+116.50$\%$) & 19.18 (+119.20$\%$) & 19.02 (+121.68$\%$) & 11.95 (+45.38$\%$) & 17.52 (+70.10$\%$) \\
				11) & SuperPoint+RANSAC & 1.74 (+13.28$\%$) & 32.29 (+97.37$\%$) & 12.97 (+48.23$\%$) & 11.83 (+37.88$\%$) & 14.90 (+81.27$\%$) & 14.75 (+43.20$\%$) \\
				12) & SuperPoint+MAGSAC & 1.86 (+33.20$\%$) & 35.19 (+115.10$\%$) & 13.34 (+52.46$\%$) & 12.90 (+50.35$\%$) & 14.28 (+73.72$\%$) & 15.51 (+50.58$\%$) \\
				13) & LoFTR+RANSAC & 1.73 (+11.62$\%$) & 16.85 (+3.00$\%$) & 25.96 (+196.69$\%$) & \underline{8.58 (+0.00$\%$)} & 10.52 (+27.98$\%$) & 12.73 (+23.59$\%$) \\
				14) & LoFTR+MAGSAC & 1.78 (+19.92$\%$) & 18.27 (+11.67$\%$) & 26.31 (+200.69$\%$) & 9.89 (+15.27$\%$) & 11.74 (+42.82$\%$) & 13.60 (+32.04$\%$) \\
				\midrule
				15) & Supervised & 94.59 (+15425.98$\%$) & 106.46 (+550.74$\%$) & 99.41 (+1036.17$\%$) & 43.40 (+405.88$\%$) & 67.13 (+716.70$\%$) & 82.20 (+698.06$\%$) \\
				16) & Unsupervised$^{*}$ & 94.13 (+15349.82$\%$) & 105.94 (+547.58$\%$) & 98.76 (+1028.69$\%$) & 43.35 (+405.22$\%$) & 67.01 (+715.26$\%$) & 81.84 (+694.55$\%$) \\
				17) & CAHomo$^{*}$ & 92.11 (+15014.96$\%$) & 99.91 (+510.70$\%$) & 91.27 (+943.14$\%$) & 35.80 (+317.24$\%$) & 58.72 (+614.33$\%$) & 75.56 (+633.63$\%$) \\
				18) & BasesHomo$^{*}$ & 77.07 (+12518.71$\%$) & 95.76 (+485.33$\%$) & 81.67 (+833.34$\%$) & 32.99 (+284.50$\%$) & 49.15 (+497.96$\%$) & 67.33 (+553.68$\%$) \\
				\midrule
				19) & UNSUPDIS & 3.49 (+303.78$\%$) & 34.00 (+107.82$\%$) & \underline{8.75 (+0.00$\%$)} & 12.29 (+43.24$\%$) & 10.93 (+32.97$\%$) & 13.89 (+34.85$\%$) \\
				20) & UNSUPDIS$^{*}$ & 3.40 (+288.84$\%$) & 34.88 (+113.20$\%$) & 9.26 (+5.83$\%$) & 11.05 (+28.79$\%$) & 9.46 (+15.09$\%$) & 13.61 (+32.14$\%$) \\
				21) & DAMG & 2.55 (+147.74$\%$) & 19.54 (+19.44$\%$) & 9.10 (+4.00$\%$) & 11.74 (+36.83$\%$) & 10.03 (+22.02$\%$) & 10.59 (+2.82$\%$) \\
				22) & DAMG$^{*}$ & 2.17 (+84.66$\%$) & 20.12 (+22.98$\%$) & 8.78 (+0.34$\%$) & 10.82 (+26.11$\%$) & 9.62 (+17.03$\%$) & \underline{10.30 (+0.00$\%$)} \\
				\midrule
				23) & Ours &  \textbf{1.66 (+0.00$\textbf{\%}$)} & \textbf{5.49 (-66.44$\textbf{\%}$)} & \textbf{4.11 (-53.03$\textbf{\%}$)} & \textbf{7.57 (-11.77$\textbf{\%}$)} & \textbf{6.95 (-15.45$\textbf{\%}$)} & \textbf{5.16 (-49.90$\textbf{\%}$)} \\
				\bottomrule
		\end{tabular}}
		\caption{The point matching errors (PME) of our method and all comparison methods on our large-baseline dataset. The best results are highlighted in bold, the second best results are underlined. The percentages in the parentheses indicate the relative change in comparison to the second best results.} 
		\label{tab:large-table}
	\end{table*}
	As shown in Table~\ref{tab:large-table}, our method achieves state-of-the-art performance in all categories of the large-baseline dataset. In the regular (RE-L) scenes, our method and SIFT+RANSAC achieve the best results, because the feature-based methods can obtain sufficient high-quality matching points in regular scenes and thus perform well. But feature-based methods fail in other challenging scenes, especially in low-light (LL-L) and low-texture (LT-L), while our method does not rely on feature detection and correspondence matching, being more robust than traditional methods in these scenarios. For example, our method reduces the error on LT-L by 66.44$\%$ compared to SIFT+MAGSAC. Moreover, image alignment methods perform better than feature-based methods in some challenging scenarios, our method still produces lower errors, e.g., our method reduces the error on LL-L by 53.03$\%$ compared to UNSUPDIS. The small-foreground (SF-L) and large-foreground (LF-L) scenes contain dynamic objects, affecting the estimation of homography. Compared with other deep learning-based methods using photometric losses for optimization, our method outperforms them in LF-L and SF-L benefiting from our homography identity loss.  
	
\begin{figure*}[!ht]
		\centering
		\includegraphics[width=0.98\linewidth]{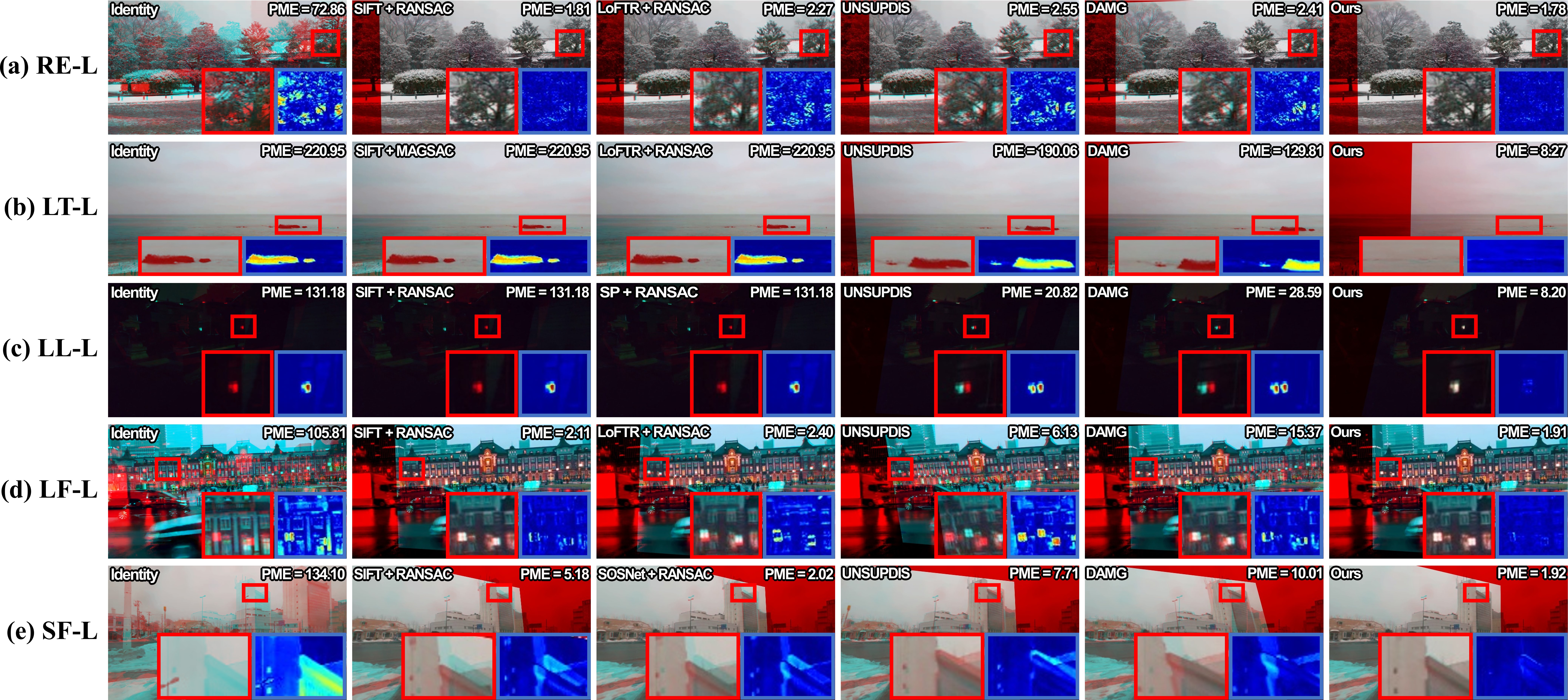}
		\caption{Qualitative results of our method and other competitive methods on the large-baseline dataset. The images are generated by superimposing the warped source images on the target image. Error-prone regions are highlighted with red boxes, and the blue boxes show the content difference between two images in the error-prone regions. Best viewed with zooming in.}
		\label{fig:large_baseline}
	\end{figure*}

	Additionally, we also conduct experiments on the small-baseline dataset~\cite{CA-Unsupervised2020}, which also contains 5 categories RE-S, LT-S, LL-S, LF-S, and SF-S. Considering the small relative motion between two images in small-baseline scenes, we set $n=1$ and reduce the non-overlap rate of the source and inserted image. As reported in Table \ref{tab:small-table}, our method outperforms the existing four deep learning-based methods with the error reduced from 0.50 to 0.44.
	\begin{table}[!t]
		\centering
		\resizebox{\linewidth}{!}{ 
			\begin{tabular}{rlllllll}
				\toprule
				1) & & RE-S & LT-S & LL-S & LF-S & SF-S & Avg-S  \\
				\midrule
				2) & $I_{3\times3}$ & 7.75 & 7.65 & 7.21 & 3.39 & 7.53 & 6.70 \\
				\hline
				3) & Supervised & 1.51 & 4.48 & 2.76 & 3.00 & 2.62 & 2.87 \\
				4) & Unsupervised & 0.79 & 2.45 & 1.48 & 1.10 & 1.11 & 1.39 \\
				5) & CAHomo & 0.73 & 1.01 & 1.03 & 0.70 & 0.92 & 0.88 \\
				6) & BasesHomo & \textbf{0.29} & \textbf{0.54} & 0.65 & 0.41 & 0.61 & 0.50 \\
				\midrule
				7) & Ours & \textbf{0.29} & 0.64 & \textbf{0.45} & \textbf{0.39} & \textbf{0.45} & \textbf{0.44} \\
				\bottomrule
		\end{tabular}}
		\caption{The point matching errors (PME) of our method and deep learning-based comparison methods on the small-baseline dataset. The best results are highlighted in bold.} 
		\label{tab:small-table}
	\end{table}
	\noindent \textbf{Qualitative Comparison.} We compare the qualitative results of our method and other methods on our large-baseline dataset. Fig.~\ref{fig:large_baseline} shows the visualization results of our method and four other methods in large-baseline scenes. Our method shows superiority in LT-L and LL-L scenes, where feature-based methods all fail due to insufficient key points extracted, as shown in Fig.~\ref{fig:large_baseline}(b) and Fig.~\ref{fig:large_baseline}(c). The other two deep image alignment methods can not align these images as well as ours. More specifically, they try to align the island in Fig.~\ref{fig:large_baseline}(b) since it has more texture than the surrounding area, while only our method successfully aligns the scene. In Fig.~\ref{fig:large_baseline}(d) and Fig.~\ref{fig:large_baseline}(e), UNSUPDIS and DAMG even perform worse than some feature-based methods because their optimization strategy is based on photo losses and thus cannot obtain satisfactory results in scenes with dynamic objects. Our method avoids the drawbacks of the photometric losses and thus generates more accurate results. Please refer to the supplementary material for more qualitative results.
	\noindent \textbf{Robustness Evaluation.} Furthermore, we evaluate the robustness of all comparison methods on the large-baseline scenes by setting thresholds to calculate the proportion of inlier predictions. Specifically, points with errors lower than the threshold are considered inliers. As such, we set the threshold from 1.0 to 50.0. Our method significantly outperforms other methods from threshold 10.0 to 50.0. With a threshold of 10, our inlier proportion is 7.7$\%$ higher than the second best (89.1$\%$ vs. 81.4$\%$), and our method does not produce any points with an error greater than 50. We have plotted a curve to reflect the robustness of our method, please refer to the supplementary material.
	\begin{table}[!t]
		\centering
		\resizebox{\linewidth}{!}{  
			\begin{tabular}{rlllllll}
				\toprule
				1) & Modification & RE-L & LT-L & LL-L & LF-L & SF-L & Avg-L  \\
				\midrule
				2) & $n=1$ & 1.91 & 6.49 & 4.71 & 7.11 & 7.26 & 5.50 \\ 
				3) & $n=3$ & 1.64 & 9.77 & 4.05 & 8.70 & 7.92 & 6.42 \\
				\midrule
				4) & Photometric loss & 59.41 & 75.47 & 53.56 & 30.91 & 48.29 & 53.53 \\ 
				5) & AB photometric loss & 5.81 & 18.56 & 15.47 & 10.62 & 10.69 & 12.23 \\ 
				6) & Only $\mathcal{L}_{sup}$ & 1.78 & 10.41 & 5.25 & 10.87 & 9.18 & 7.50  \\ 
				7) & Only $\mathcal{L}_{unsup}$ & 2.50 & 8.63 & 5.84 & 6.78 & 7.07 & 6.16 \\ 
				\midrule
				8) & Corner offsets  & 4.73 & 13.21 & 9.31 & 15.19 & 15.09 & 11.51 \\ 
				9) & 8 bases flow & 7.14 & 13.04 & 13.18 & 12.50 & 14.56 & 12.09 \\ 
				\midrule
				10) & Default &  1.66 & 5.49 & 4.11 & 7.57 & 6.95 & 5.16 \\
				\bottomrule
		\end{tabular}}
		\caption{Results of ablation studies, please refer to the text for more details.}
		\label{tab:ablation}
	\end{table}
	\subsection{Ablation Studies}
	We conduct extensive ablation studies to verify the effectiveness of our proposed components, and the results are reported in Table~\ref{tab:ablation}.
	
	\noindent \textbf{Progressive Estimation Strategy.} In this experiment, we choose to vary the number of inserted intermediate images to verify the effectiveness of our progressive estimation strategy, as shown in rows 2 and 3. With only one image inserted, i.e., $n=1$, the average error of our method is 5.50, which already surpasses other comparison methods in Table~\ref{tab:large-table}, but can still be reduced. After inserting three images, i.e., $n=3$, the average error increased by 1.26 compared to $n=2$. This is because the more images are inserted, the more accurate intermediate results should be interpolated, otherwise, the cumulative error increases significantly~\cite{content-stabilization,stabilization}. Even though the number of inserted images varies, our strategy still achieves superior results compared to other methods, and the optimal results can be obtained when the number of inserted images is moderate.
	
	\noindent \textbf{Homography Identity Loss.} To verify the effectiveness of our homography identity loss, we first compare our loss function with the photometric losses used in~\cite{BasesHomo2021} and~\cite{stitching1}, as shown in rows 4 and 5. We can see that optimizing with photometric loss~\cite{BasesHomo2021} leads to failure in all scenarios, which is consistent with what we have discussed in the Introduction. The ablation-based photometric loss(AB photometric loss)~\cite{stitching1} avoids the effect of out-of-boundary pixels, it is not able to handle scenes with dynamic objects. Our proposed loss performs better than both of them in all scenes, which demonstrates the usefulness of our loss function. In addition, our homography identity loss is a semi-supervised loss, to verify the effectiveness of semi-supervised learning, we remove the $\mathcal{L}_{sup}$ and $\mathcal{L}_{unsup}$ respectively for optimization. As shown in rows 6 and 7, the average error of training with $\mathcal{L}_{sup}$ only is higher than that of training with $\mathcal{L}_{unsup}$ only (7.50 vs. 6.16), which is due to the fact that our supervised objective is constructed based on the synthetic data and is therefore not ideal in terms of generalizability. But solely using $\mathcal{L}_{unsup}$ is worse than solely using $\mathcal{L}_{sup}$ in RE-L and LL-L scenarios. By combining the advantages of supervised and unsupervised learning, our semi-supervised loss can achieve better results. 
	
	\noindent \textbf{Homography Flow.} In the training stage, we adopt the homography flow to facilitate network training. From another perspective, our homography flow is equivalent to a form of dense offsets, similar to the commonly used corner offsets~\cite{unsupervised2018,CA-Unsupervised2020}. However, our homography flow contains more motion information and achieves better results, as shown in row 8. Another similar form is the 8 bases flow~\cite{BasesHomo2021}, it performs well in the small-baseline scenes but crashes in the large-baseline, as shown in row 9.
	
	\subsection{Limitations}
	Although our method achieves state-of-the-art performance in large-baseline scenes compared with the existing methods, it still has its limitation of being applied to scenes with multiple planes where a homography theoretically cannot perform alignment well. We will leave the solution for the multiple planes alignment as future work. 
	
	\section{Conclusion}
	In this work, we have presented a new deep framework for large-baseline homography estimation. We note that it is non-trivial to directly estimate a large-baseline homography and thus propose a progressive estimation strategy to convert it into several intermediate phases. The homography of two images can be obtained by cumulatively multiplying these intermediate ones. Meanwhile, we propose a semi-supervised homography identity loss to enforce the network focus on optimizing the homography itself, avoiding the problems of photometric losses. Moreover, we present a large-scale dataset for large-baseline homography estimation, which consists of five categories of scenes. Extensive experiments and ablation studies prove the effectiveness of our newly proposed components and demonstrate the superiority of our method over the existing methods.
	
	\section{Acknowledgements}
	This work was supported by the National Natural Science Foundation of China (NSFC) under grant No.61872067.

	\bibliography{LBHomo}
\end{document}